\documentclass[10pt,twocolumn,letterpaper]{article}
\usepackage{cvpr}
\usepackage{graphicx}
\usepackage{amsmath}
\usepackage{amssymb}
\usepackage{booktabs}
\usepackage{lipsum}
\usepackage{relsize}
\usepackage{subcaption}
\usepackage{multicol}
\makeatletter
\@namedef{ver@everyshi.sty}{}
\makeatother
\usepackage{pgfplots}
\pgfplotsset{compat=1.7}

\usepackage[pagebackref,breaklinks,colorlinks]{hyperref}

\usepackage[capitalize]{cleveref}
\crefname{section}{Sec.}{Secs.}
\Crefname{section}{Section}{Sections}
\Crefname{table}{Table}{Tables}
\crefname{table}{Tab.}{Tabs.}

\def\cvprPaperID{*****} 
\def\confName{CVPR}
\def\confYear{2022}

\begin{document}

\title{Adversarially Robust Medical Classification via Attentive Convolutional Neural Networks}

\author{Isaac Wasserman\\
  University of Pennsylvania\\
  {\tt\small isaacrw@seas.upenn.edu}
}
\maketitle

\begin{abstract}
  Convolutional neural network-based medical image classifiers have been shown to be especially susceptible to adversarial examples. Such instabilities are likely to be unacceptable in the future of automated diagnoses. Though statistical adversarial example detection methods have proven to be effective defense mechanisms, additional research is necessary that investigates the fundamental vulnerabilities of deep-learning-based systems and how best to build models that jointly maximize traditional and robust accuracy. This paper presents the inclusion of attention mechanisms in CNN-based medical image classifiers as a reliable and effective strategy for increasing robust accuracy without sacrifice. This method is able to increase robust accuracy by up to 16\% in typical adversarial scenarios and up to 2700\% in extreme cases. 
\end{abstract}

\section{Introduction}
For years, neural networks have been applied to medical diagnostic tasks and have achieved levels of performance on par with highly trained pathologists, radiologists, and other diagnosticians \cite{medical-cnn-survey}. However, these systems are often relegated to lab settings and clinical decision support systems, as their black-box nature does not inspire the confidence necessary for life-critical environments \cite{NIH-AI}\cite{AI-CDSS}.
    
The decisions of such models are near impossible for the researchers that created them to understand, let alone the proposed clinical end-user \cite{Med-XAI}. Additionally, neural networks are incredibly vulnerable to adversarial examples. Adversarial examples are defined as instances, $x$, whose true label is certifiably $y$, but the mechanics of the estimator (either inherent to the architecture or specific to the weights) result in a confidently incorrect prediction \cite{Intro-Adv}. These examples can be carefully contrived instances, $x^\prime = x + p$, which are extremely similar to a natural instance, $x$, but are classified differently, $h(x) \not = h(x^\prime)$. However, adversarial examples aren't just the product of bad-actors. Recent research has demonstrated the existence of many naturally occurring instances, which behave adversarially when used as input to popular models \cite{Natural-Adv}. Prior to this discovery, researchers and engineers building models for environments where security and bad-actors were not a concern could somewhat understandably deprioritize adversarial robustness, but with the combination of increased reliance on AI-driven systems, the existence of natural adversarial examples, and the proliferation of civilian-targeted cyber-warfare, secure and reliable neural networks are more important than ever.

For medical applications, clean- and robust-accuracy must be optimized simultaneously and with near-equal priority. If the goal of medical AI is to increase accessibility to high-quality diagnostics and treatments, humans will need to abdicate their roles in some procedural medical processes where AI succeeds, such as image classification and segmentation. Unfortunately, these are also the applications where a well-placed orthopedic pin or speck of dust could make all the difference between a properly and improperly diagnosed patient.

This paper proposes a simple architectural suggestion for medical image classification models that greatly increases robust-accuracy without sacrificing clean-accuracy (i.e. accuracy on non-adversarial inputs), unlike previous techniques \cite{RobustVsAccuracy}. Additionally, this method increases the parameter count of ResNet-50-based \cite{ResNet} architectures by less than 1.3\%; therefore, training time and data requirements are not significantly affected. Later sections will carefully compare the behavioral differences between baseline and adjusted architectures to investigate the root-cause of this increased performance.

\section{Related Work}
  \subsection{Vulnerability of Medical Image Models}
    Paschali et al. (2018) was among the first to examine the susceptibility of medical imaging models to adversarial examples, discovering that current attacks were able to decrease accuracy on popular medical classification models by up to 25\% and medical segmentation models by up to 41\% \cite{Paschali}. Building on this work, as well as that of Finlayson et al. (2019) \cite{Finlayson}, Ma and Niu et al. (2021) found that models designed for medical images were, in fact, more vulnerable to adversarial image attacks than those designed for natural images. Key to this conclusion was their finding that the medical image models they tested had significantly sharper loss landscapes than their natural image counterparts \cite{MaNiu}. This correlation between sharp loss landscapes and more vulnerable models is outlined by Madry et al. (2017), which attributes this sharpness to overparametrization \cite{Madry}. Ma and Niu et al. (2021) echoes this concern of overcomplexity and also suspects that the salience of intricate textures in medical images may also contribute to their compatibility with adversarial attacks. Additionally, they find that while medical image models are easily fooled by adversarial images, adversarially perturbed medical images are more easily detected than their natural image counterparts; they attribute this property to the tendency of popular attacks to place perturbations outside of the image's salient region \cite{MaNiu}.

  \subsection{Finding Adversarially Robust Architectures}
    Training adversarially robust neural networks is an extremely active area of research. Current best-practices involve adversarial training, in which adversarial examples are included in the training set \cite{Madry}. However, this method serves neither to remedy nor understand the underlying vulnerabilities of neural networks and often comes at the cost of training time and clean accuracy \cite{RobustVsAccuracy}. For this reason, it is imperative that the research community continues to investigate architectural modifications that yield greater adversarial robustness.

    Though many recent works have developed neural architecture search algorithms for identifying robust architectures \cite{AdvRush}\cite{MORAS}\cite{DSRNA}, these methods are merely choosing the most robust architecture from a finite, user-defined search space. This is a computationally expensive process that requires a large amount of labeled data. Additionally, the architectures found are specific to a single dataset, and the robustness does not generalize to other datasets.

  \subsection{CNNs with Attention}
    Recent advances in natural language processing have offered the computer vision research community a new option for image classification, the vision transformer (ViT) \cite{VisionTransformerSurvey}. Models utilizing this family of architectures commonly match or exceed the performance of convolutional neural networks \cite{ViT}\cite{DeiT}\cite{PVT}\cite{Swin}. Shao et al. (2021) found that ViTs tend to learn more robust, high-level features, allowing them to ignore the high frequency perturbations of many attack methods. When compared to various versions of ResNet, ShuffleNet, MobileNet, and VGG16, ViT-S/16 was up to 46\% more robust to PGD attacks and 44\% more robust to AutoAttack \cite{RobustTransformers}.

    Since their inception, it has been understood that the superior clean-accuracy of vision transformers is related to their reliance on attention \cite{VisionTransformerSurvey}, and recently it has been confirmed that this property is also responsible for the architecture's superior robustness \cite{UnderstandingTransformerRobustness}. Zhou et al. (2022) found that the self-attention of vision transformers tends to promote the saliency of more meaningful (non-spurious) clusters of image regions.

    Working from this knowledge, it is natural to wonder whether attention mechanisms can offer additional adversarial robustness to CNNs. Agrawal et al. (2022) concluded that this was not necessarily the case, finding that while their attentive CNNs had slightly superior robustness to PGD attacks on the CIFAR-100 dataset, it fell behind ResNet-50 on CIFAR-10 and Fashion MNIST. Based on these results, they suspected that the adversarial robustness of attentive models on a given dataset may correlate to the number of classes \cite{AttentiveCNNRobustness}.

    Additionally, early research on CNNs with attention by Xiao et al. (2015) found that attentive CNNs had significantly higher clean-accuracy for fine-grained classification than vanilla-CNNs. As small details are incredibly salient in medical image datasets, research on fine-grained classification is especially relevant. Based on these findings and those concerning the role of attention in robust-accuracy, it appears possible that the use of attention is a minimal architectural feature that challenges the findings of Tsipras et al. (2018) \cite{RobustVsAccuracy} by improving clean- and robust-accuracy simultaneously.

  \section{Method}
    \subsection{Datasets}
    Benchmark medical image classification tasks were chosen based on their frequent use in similar studies, such as Ma and Niu et al. (2021) \cite{MaNiu} and Finlayson et al. (2019) \cite{Finlayson}. These tasks are diabetic retinopathy detection from fundoscopy images, pneumothorax detection from chest x-rays, and skin lesion classification from dermatoscopy images. Fundoscopy images were sourced from the popular Kaggle diabetic retinopathy detection competition \cite{KaggleDR}. Chest x-rays used were from the ChestX-Ray14 dataset \cite{ChestX-Ray14}.

    \subsection{Network Architecture}
      Four models were trained for each dataset, a ResNet-50 with and without a soft attention block and an InceptionResNetV2 with and without soft attention. The architecture of each model was based on the TensorFlow \cite{TensorFlow} implementations of ResNet-50 \cite{ResNetImplementation} and InceptionResNetV2 \cite{InceptionImplementation}.
      
      ResNet-50 was instantiated with the top block included, initial weights based on ImageNet \cite{ImageNet} pretraining, and class count set to 1000. The final three layers of the network were removed. In the models without attention, these final layers were replaced with a 2D global average pooling layer followed by a fully-connected layer with softmax activation. In the models with attention, the last three layers were replaced with a soft attention block implemented according to the specifications of Datta et al. (2021). This block produces a $7 \times 7$ feature map which is $2 \times 2$ max pooled and concatenated with a $2 \times 2$ max pooled version of the input to the attention block. This concatenated output is put through ReLU activation, 50\% dropout, and global average pooling before being handed off to the fully connected prediction head with softmax activation \cite{AttentionSkinCancerClassification}.
      
      InceptionResNetV2 \cite{InceptionResNet} was instantiated with the top block included, initial weights based on ImageNet, and classifier activation as softmax. In the models without attention, the final 28 layers were replaced with a ReLU activation and 50\% dropout whose output was flattened and sent to the fully connected prediction head with softmax activation. In the models with attention, these layers were replaced by the soft attention block described above. This output is $2 \times 2$ max pooled and concatenated with a max pooled version of the input to the attention block before being sent through ReLU activation and the softmax prediction head.

      All models used the Adam optimizer with $\eta=0.01$ and $\epsilon=0.1$ and minimized categorical cross-entropy during training. During training classes were weighted based on their inverse frequency relative to the other classes. Each model was trained for a maximum of 300 epochs with early stopping (patience $=$ 40, minimum delta $=$ 0.001) causing most to stop after 60-90 epochs. In the dermatoscopy task, models only saw 10\% of the dataset during each epoch; this was done to increase the odds of reproducing the results of Datta et al. (2021) \cite{AttentionSkinCancerClassification}.

    \subsection{Evaluation and Analysis of Robustness}
      After training, the models' clean- and robust-accuracy was evaluated using the FoolBox \cite{FoolBoxPaper}\cite{FoolBoxLibrary} library's implementation of $l_\infty$ projected gradient descent attacks. Each set of models was tested at increasing epsilons (perturbation radii). Attacks of these perturbation radii were created for each image in the test sets. Unweighted accuracy was calculated for each perturbation radius.

      \begin{figure}[h]
    \begin{subfigure}{\linewidth}
      \centering
      $
      \begin{array}{l}
      \includegraphics[width=0.26\linewidth]{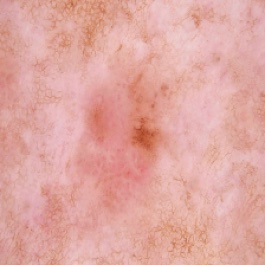}
      \end{array}
      \mathlarger{\mathlarger{\mathlarger{\mathlarger{+}}}}
      \begin{array}{l}
        \includegraphics[width=0.26\linewidth]{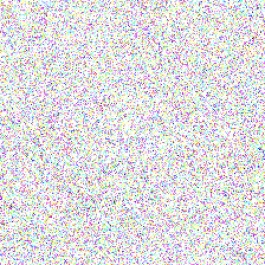}
      \end{array}
      \mathlarger{\mathlarger{\mathlarger{\mathlarger{=}}}}
      \begin{array}{l}
        \includegraphics[width=0.26\linewidth]{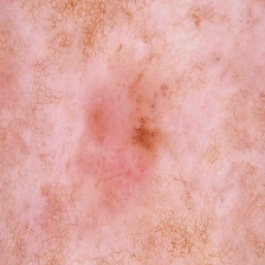}
      \end{array}
      $
      \caption{An example of $\epsilon=0.01$ perturbation on the dermatoscopy dataset. 0.01 is the maximum perturbation radius used for this dataset.\\}
      \label{example_derm_perturbation}
    \end{subfigure}
    
    \begin{subfigure}{\linewidth}
      \centering
      $
      \begin{array}{l}
      \includegraphics[width=0.26\linewidth]{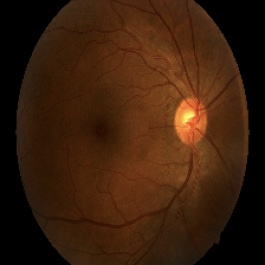}
      \end{array}
      \mathlarger{\mathlarger{\mathlarger{\mathlarger{+}}}}
      \begin{array}{l}
        \includegraphics[width=0.26\linewidth]{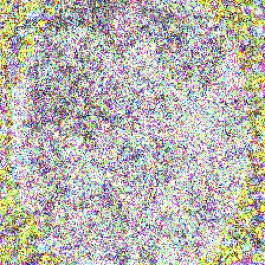}
      \end{array}
      \mathlarger{\mathlarger{\mathlarger{\mathlarger{=}}}}
      \begin{array}{l}
        \includegraphics[width=0.26\linewidth]{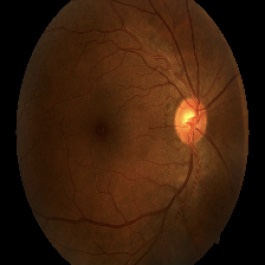}
      \end{array}
      $
      \caption{An example of $\epsilon=0.32$ perturbation on the fundoscopy dataset. 0.32 is the maximum perturbation radius used for this dataset.\\}
      \label{example_dr_perturbation}
    \end{subfigure}
    \begin{subfigure}{\linewidth}
      \centering
      $
      \begin{array}{l}
      \includegraphics[width=0.26\linewidth]{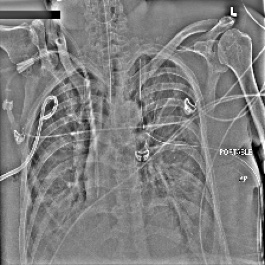}
      \end{array}
      \mathlarger{\mathlarger{\mathlarger{\mathlarger{+}}}}
      \begin{array}{l}
        \includegraphics[width=0.26\linewidth]{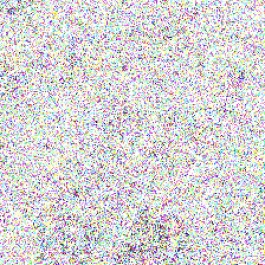}
      \end{array}
      \mathlarger{\mathlarger{\mathlarger{\mathlarger{=}}}}
      \begin{array}{l}
        \includegraphics[width=0.26\linewidth]{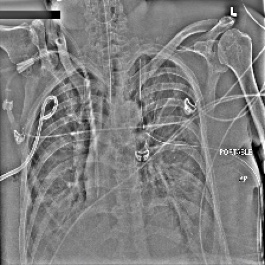}
      \end{array}
      $
      \caption{An example of $\epsilon=0.32$ perturbation on the chest x-ray dataset. 0.32 is the maximum perturbation radius used for this dataset.\\}
      \label{example_pneumo_perturbation}
    \end{subfigure}%
    \caption{Perturbation examples}
  \end{figure}

  \section{Results}
    \subsection{ResNet-50 Models}
      \begin{figure}[h]
    \begin{tikzpicture}
      \begin{axis}[xtick={0, 0.00125, 0.0025, 0.005, 0.01, 0.02, 0.04, 0.08, 0.16, 0.32}, x tick label style={rotate=45, log ticks with fixed point},xmode=log, log basis x=2, xlabel=Perturbation Radius ($\epsilon$), ylabel=Accuracy, width=\linewidth, height=7cm,legend style={at={(0.95,0.95)},anchor=north east}]
          
      \addplot[color=red,mark=x] coordinates {
        (0.00125, 0.687)
        (0.0025, 0.216)
        (0.005, 0.012)
        (0.01, 0)
      };
      
      \addplot[color=blue,mark=*] coordinates {
        (0.00125, 0.795)
        (0.0025, 0.685)
        (0.005, 0.337)
        (0.01, 0.022)
      };

      \addplot[color=red, domain=0.00125:0.01, dashed]{0.905};
      \addplot[color=blue, domain=0.00125:0.01, dashed]{0.880};
      
      \legend{Without Attention,With Attention}
      \end{axis}
      \end{tikzpicture}
    \caption{Clean- and robust-accuracy of ResNet-50 models for skin lesion classification.  Dashed lines represent clean-accuracy.}
    \label{DermResNet50Robustness}
  \end{figure}
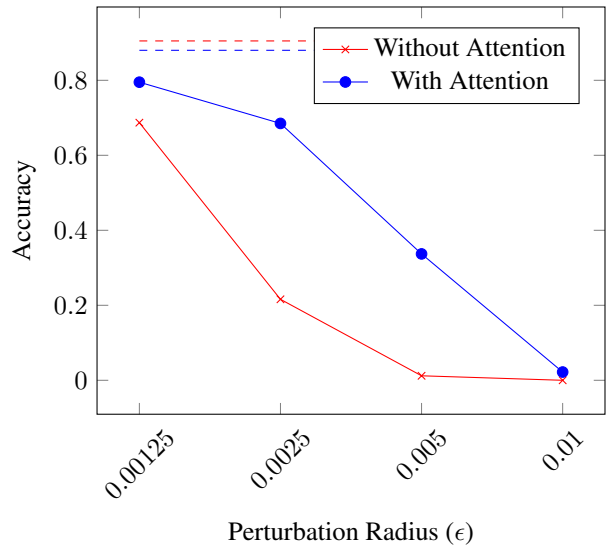
      \begin{figure}[h]
    \begin{tikzpicture}
      \begin{axis}[xtick={0, 0.00125, 0.0025, 0.005, 0.01, 0.02, 0.04, 0.08, 0.16, 0.32}, x tick label style={rotate=45, log ticks with fixed point},xmode=log, log basis x=2, xlabel=Perturbation Radius ($\epsilon$), ylabel=Accuracy, width=\linewidth, height=7cm,legend style={at={(0.05,0.05)},anchor=south west}]
          
      \addplot[color=red,mark=x] coordinates {
        (0, 0.832)
        (0.00125, 0.828)
        (0.0025, 0.825)
        (0.005, 0.819)
        (0.01, 0.806)
        (0.02, 0.778)
        (0.04, 0.705)
        (0.08, 0.506)
        (0.16, 0.175)
        (0.32, 0.014)
      };
      
      \addplot[color=blue,mark=*] coordinates {
        (0, 0.818)
        (0.00125, 0.816)
        (0.0025, 0.815)
        (0.005, 0.813)
        (0.01, 0.808)
        (0.02, 0.8)
        (0.04, 0.780)
        (0.08, 0.733)
        (0.16, 0.599)
        (0.32, 0.287)
      };

      \addplot[color=red, domain=0.00125:0.32, dashed]{0.832};
      \addplot[color=blue, domain=0.00125:0.32, dashed]{0.818};
      
      \legend{Without Attention,With Attention}
      \end{axis}
      \end{tikzpicture}
    \caption{Clean- and robust-accuracy of ResNet-50 models for diabetic retinopathy detection. Dashed lines represent clean-accuracy.}
    \label{DRResNet50Robustness}
  \end{figure}
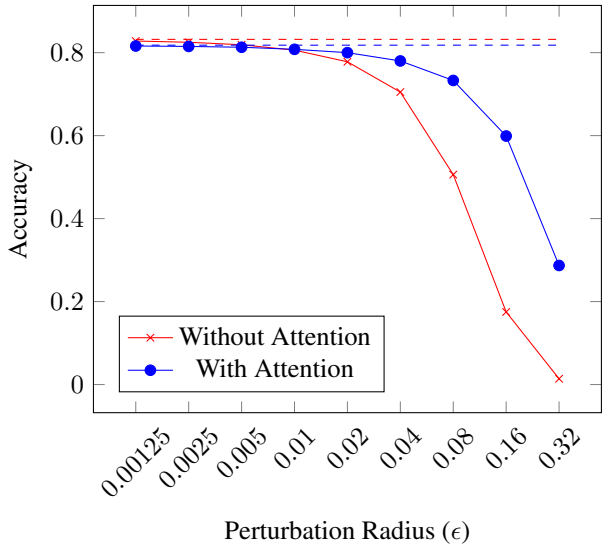
      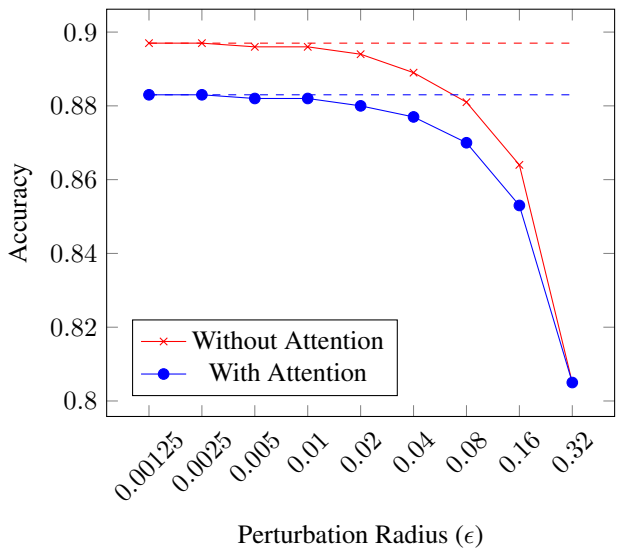
\begin{figure}[h]
    \begin{tikzpicture}
      \begin{axis}[xtick={0, 0.00125, 0.0025, 0.005, 0.01, 0.02, 0.04, 0.08, 0.16, 0.32}, x tick label style={rotate=45, log ticks with fixed point},xmode=log, log basis x=2, xlabel=Perturbation Radius ($\epsilon$), ylabel=Accuracy, width=\linewidth, height=7cm,legend style={at={(0.05,0.05)},anchor=south west}]
          
      \addplot[color=red,mark=x] coordinates {
        (0.00125, 0.897)
        (0.0025, 0.897)
        (0.005, 0.896)
        (0.01, 0.896)
        (0.02, 0.894)
        (0.04, 0.889)
        (0.08, 0.881)
        (0.16, 0.864)
        (0.32, 0.805)
      };
      
      \addplot[color=blue,mark=*] coordinates {
        (0.00125, 0.883)
        (0.0025, 0.883)
        (0.005, 0.882)
        (0.01, 0.882)
        (0.02, 0.880)
        (0.04, 0.877)
        (0.08, 0.870)
        (0.16, 0.853)
        (0.32, 0.805)
      };

      \addplot[color=red, domain=0.00125:0.32, dashed]{0.897};
      \addplot[color=blue, domain=0.00125:0.32, dashed]{0.883};
      
      \legend{Without Attention,With Attention}
      \end{axis}
      \end{tikzpicture}
    \caption{Clean- and robust-accuracy of ResNet-50 models for pneumothorax detection. Dashed lines represent clean-accuracy.}
    \label{PneumoResNet50Robustness}
  \end{figure}

      For the task of skin lesion classification, the models with attention are clearly superior, in terms of robustness (Fig. \ref{DermResNet50Robustness}). Although the model without attention has slightly higher clean-accuracy (0.905 vs. 0.88), even the slightest perturbation ($\epsilon=0.00125$) is able to reduce its accuracy by 24\% and put the model with attention in the lead. By the time $\epsilon=0.005$, the model is worse than random, and by the time $\epsilon=0.01$, the model is 0\% accurate. It is worth noting that perturbation of this size (Fig. \ref{example_derm_perturbation}) is far from being human perceptible. Meanwhile, at this perturbation radius, the model with attention remains better than random selection.

      The models for diabetic retinopathy detection tell a slightly different story (Fig. \ref{DRResNet50Robustness}). These models were, overall, much more robust, requiring a perturbation of at least $\epsilon=0.01$ to reduce accuracy by 3\%. As in the skin lesion classification models, the model without attention has a slightly higher clean accuracy (0.832 vs. 0.818). It retains this very small lead until $\epsilon=0.02$, at which point, the accuracy of both models begins falling, but the model with attention does so a bit less dramatically. At the maximum perturbation radius tested ($\epsilon=0.32$) (Fig. \ref{example_dr_perturbation}), the accuracy of the model with attention is over 20 times higher than that of the model without.

      The pneumothorax detection models, collectively, were even more robust than the diabetic retinopathy detection models (Fig. \ref{PneumoResNet50Robustness}). After perturbations of $\epsilon=0.32$ (Fig. \ref{example_pneumo_perturbation}), the accuracy of both models met at 0.805. Prior to this, the accuracy of the model without attention hovered 1-2\% above that of the model with attention. However, based on the trajectory of the models' accuracy with respect to $\epsilon$, perturbation radii higher than 0.32 would likely result in the model with attention leapfrogging the model without.

    \subsection{InceptionResNetV2 Models}

      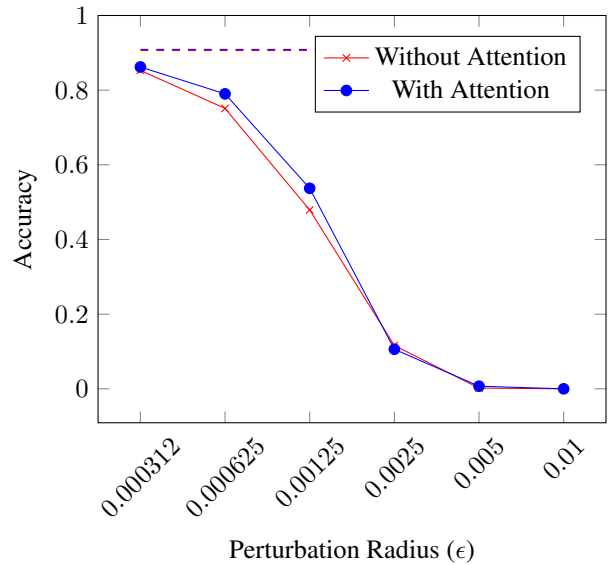
\begin{figure}[h]
    \begin{tikzpicture}
      \begin{axis}[xtick={0, 0.0003125, 0.000625, 0.00125, 0.0025, 0.005, 0.01}, x tick label style={rotate=45, log ticks with fixed point},xmode=log, log basis x=2, xlabel=Perturbation Radius ($\epsilon$), ylabel=Accuracy, width=\linewidth, height=7cm,legend style={at={(0.95,0.95)},anchor=north east}]
          
      \addplot[color=red,mark=x] coordinates {
        (0.0003125, 0.853)
        (0.000625, 0.751)
        (0.00125, 0.479)
        (0.0025, 0.116)
        (0.005, 0.002)
        (0.01, 0.0)
      };
      
      \addplot[color=blue,mark=*] coordinates {
        (0.0003125, 0.862)
        (0.000625, 0.790)
        (0.00125, 0.537)
        (0.0025, 0.106)
        (0.005, 0.007)
        (0.01, 0.0)
      };

      \addplot[color=red, domain=0.0003125:0.01, dashed]{0.906};
      \addplot[color=blue, domain=0.0003125:0.01, dashed]{0.909};
      
      \legend{Without Attention,With Attention}
      \end{axis}
      \end{tikzpicture}
    \caption{Clean- and robust-accuracy of InceptionResNetV2 models for skin lesion classification. Dashed lines represent clean-accuracy.}
    \label{IRV2DermRobustness}
  \end{figure}
      \begin{figure}[h]
    \begin{tikzpicture}
      \begin{axis}[xtick={0, 0.0003125, 0.000625, 0.00125, 0.0025, 0.005, 0.01}, x tick label style={rotate=45, log ticks with fixed point},xmode=log, log basis x=2, xlabel=Perturbation Radius ($\epsilon$), ylabel=Accuracy, width=\linewidth, height=7cm,legend style={at={(0.95,0.95)},anchor=north east}]
      
      \addplot[color=red,mark=x] coordinates {
        (0.0003125, 0.618)
        (0.000625, 0.286)
        (0.00125, 0.053)
        (0.0025, 0.007)
        (0.005, 0.0)
        (0.01, 0.0)
      };
      
      \addplot[color=blue,mark=*] coordinates {
        (0.0003125, 0.590)
        (0.000625, 0.255)
        (0.00125, 0.046)
        (0.0025, 0.0)
        (0.005, 0.0)
        (0.01, 0.0)
      };

      \addplot[color=red, domain=0.0003125:0.01, dashed]{0.889};
      \addplot[color=blue, domain=0.0003125:0.01, dashed]{0.887};
      
      \legend{Without Attention,With Attention}
      \end{axis}
      \end{tikzpicture}
    \caption{Clean- and robust-accuracy of InceptionResNetV2 models for diabetic retinopathy detection. Dashed lines represent clean-accuracy.}
    \label{IRV2DRRobustness}
  \end{figure}
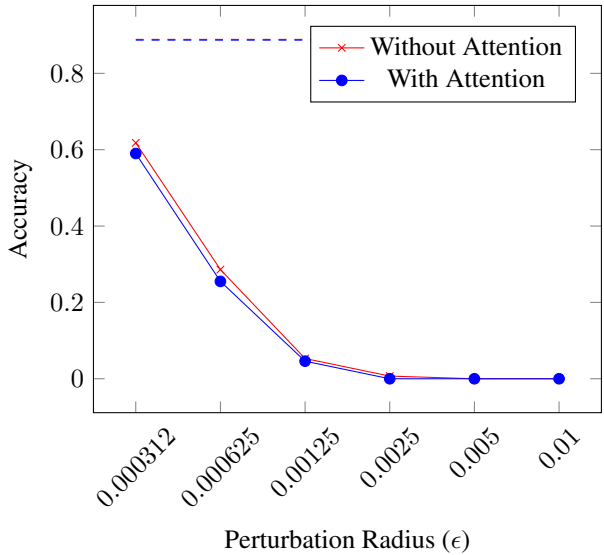
      \begin{figure}[h]
    \begin{tikzpicture}
      \begin{axis}[xtick={0, 0.0003125, 0.000625, 0.00125, 0.0025, 0.005, 0.01}, x tick label style={rotate=45, log ticks with fixed point},xmode=log, log basis x=2, xlabel=Perturbation Radius ($\epsilon$), ylabel=Accuracy, width=\linewidth, height=7cm,legend style={at={(0.95,0.95)},anchor=north east}]

      \addplot[color=red,mark=x] coordinates {
        (0.0003125, 0.984)
        (0.000625, 0.982)
        (0.00125, 0.969)
        (0.0025, 0.945)
        (0.005, 0.740)
        (0.01, 0.137)
      };
      
      \addplot[color=blue,mark=*] coordinates {
        (0.0003125, 0.972)
        (0.000625, 0.949)
        (0.00125, 0.817)
        (0.0025, 0.334)
        (0.005, 0.334)
        (0.01, 0.0)
      };

      \addplot[color=red, domain=0.0003125:0.01, dashed]{0.986};
      \addplot[color=blue, domain=0.0003125:0.01, dashed]{0.987};
      
      \legend{Without Attention,With Attention}
      \end{axis}
      \end{tikzpicture}
    \caption{Clean- and robust-accuracy of InceptionResNetV2 models for pneumothorax detection. Dashed lines represent clean-accuracy.}
    \label{IRV2PneumoRobustness}
  \end{figure}
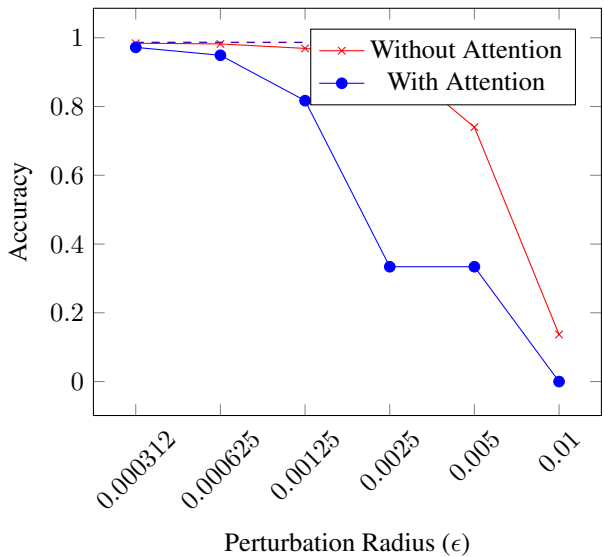

      Experiments with the InceptionResNetV2 architecture paint a slightly different picture. In all tasks, clean-accuracy for the baseline and attentive models was within 1\%. This result is, somewhat, at odds with those of Datta et al. (2021) \cite{AttentionSkinCancerClassification} which found soft attention to improve accuracy of InceptionResNetV2 (on HAM10000 \cite{HAM10000}) by ~3\%.

      For the skin lesion classification and diabetic retinopathy detection tasks, the baseline and attentive architectures performed near-identically in adversarial scenarios. The diabetic retinopathy task slightly favored the baseline, whereas skin lesion classification slightly favored the attentive model; however, at most, there was a 6\% discrepancy between their accuracies ($\epsilon = 0.00125$).
      
      In the pneumothorax detection task, the model with attention performed significantly worse under adversarial scenarios. The model reached a level of accuracy worse than random selection after the perturbation radius was pushed beyond 0.00125, while the baseline remained usable with a perturbation radius of 0.005.

  \section{Discussion}
    \subsection{Analysis of Perturbations and Activation Maps}
    In an attempt to better understand the principles and behaviors that led the CNNs with attention to perform better in both clean and adversarial scenarios, perturbation difference maps and Grad-CAM activation maps were generated for a select sample of images on each model. Individually, these difference maps and activation maps were unremarkable. However, a number of dataset specific patterns were noticed throughout the images used for this analysis.
    \subsubsection{ResNet-50 Models}
\begin{figure}[h]
    \centering
    \begin{subfigure}{.4\linewidth}
      \centering
      \includegraphics[width=\linewidth]{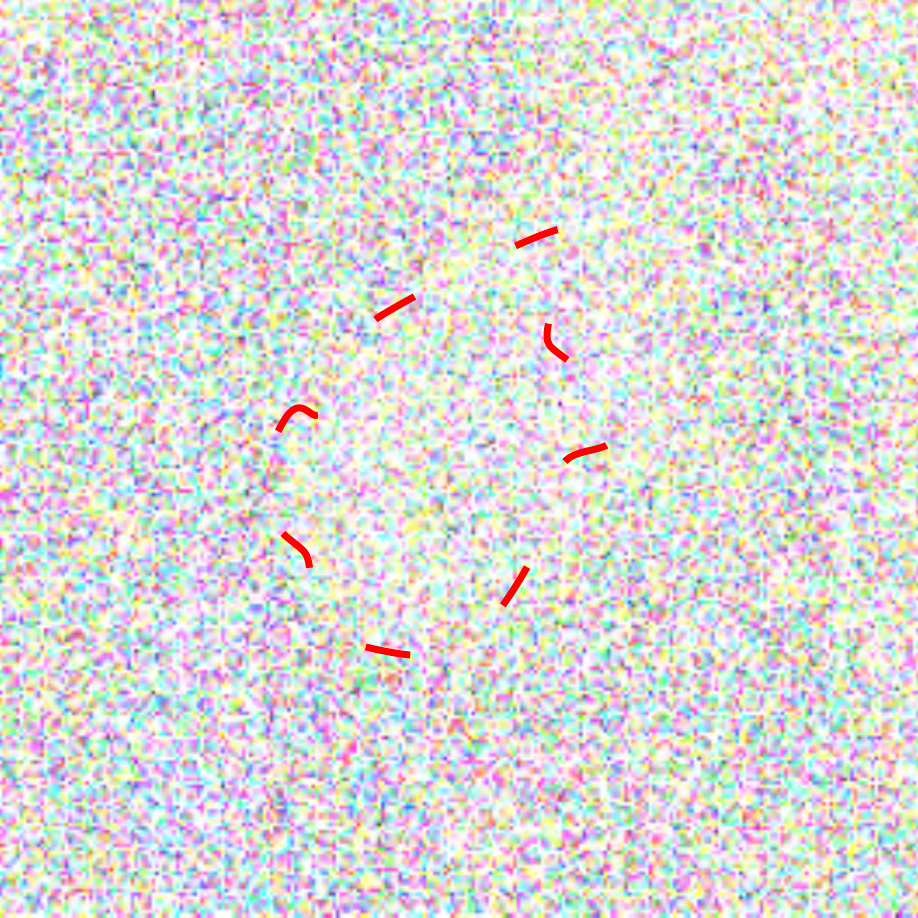}
      \caption{Shape of skin lesion found in adversarial ``noise''\\}
      \label{LesionShape}
    \end{subfigure}
    \begin{subfigure}{.4\linewidth}
      \centering
      \includegraphics[width=\linewidth]{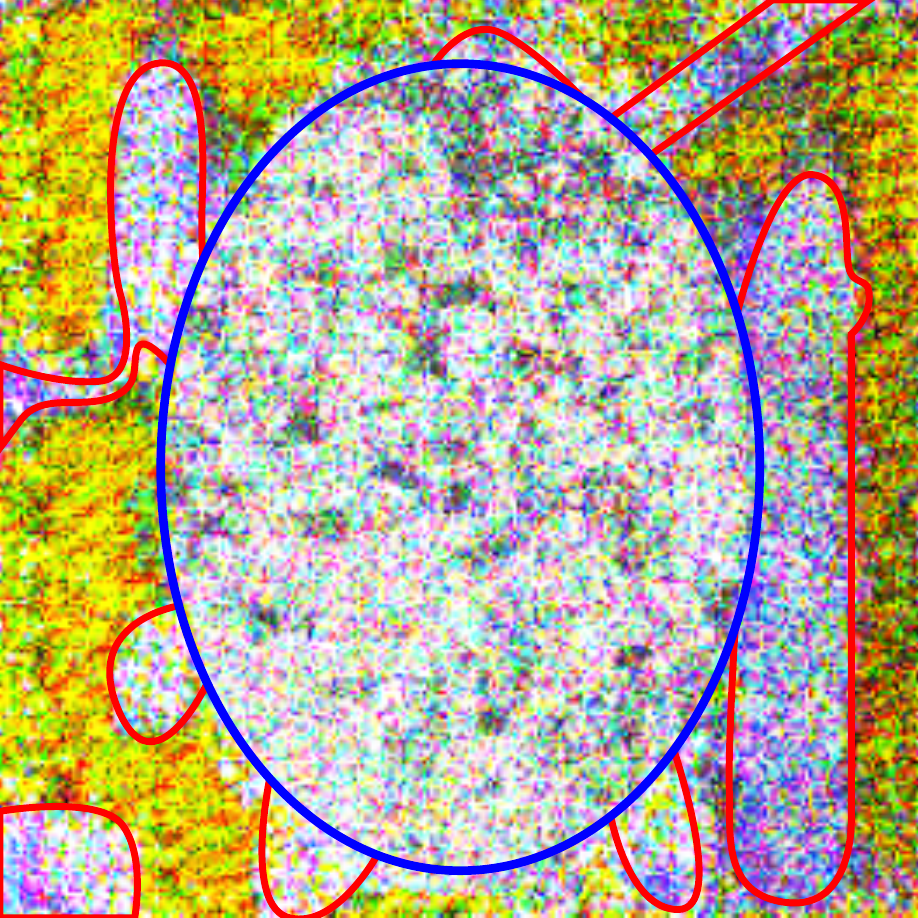}
      \caption{Divergence from shape of eye\\}
      \label{EyeShapeDivergence}
    \end{subfigure}
    \caption{Shape information carried over into perturbations}
  \end{figure}

      While imperceptible in the final images, the perturbations for dermatoscopic and fundoscopic images were found to carry easily perceptible information about the source image, when the changes were scaled to a range of $[0,255]$. In these cases, the adversarial ``noise'' contained the shape of the lesion or eye. For dermatoscopic images, this shape was in the form of a densely perturbed ring enclosing a sparsely perturbed center  (Fig. \ref{LesionShape}). This phenomenon was most visible when $\epsilon=0.00125$. For the fundoscopic images, this shape was represented by a sparsely perturbed ellipse. In the attacks generated for the model with attention, the shape of this ellipse occasionally diverged from the true shape of the eye  (Fig. \ref{EyeShapeDivergence}).

      \begin{figure}[h]
    \centering
    \includegraphics[width=\linewidth]{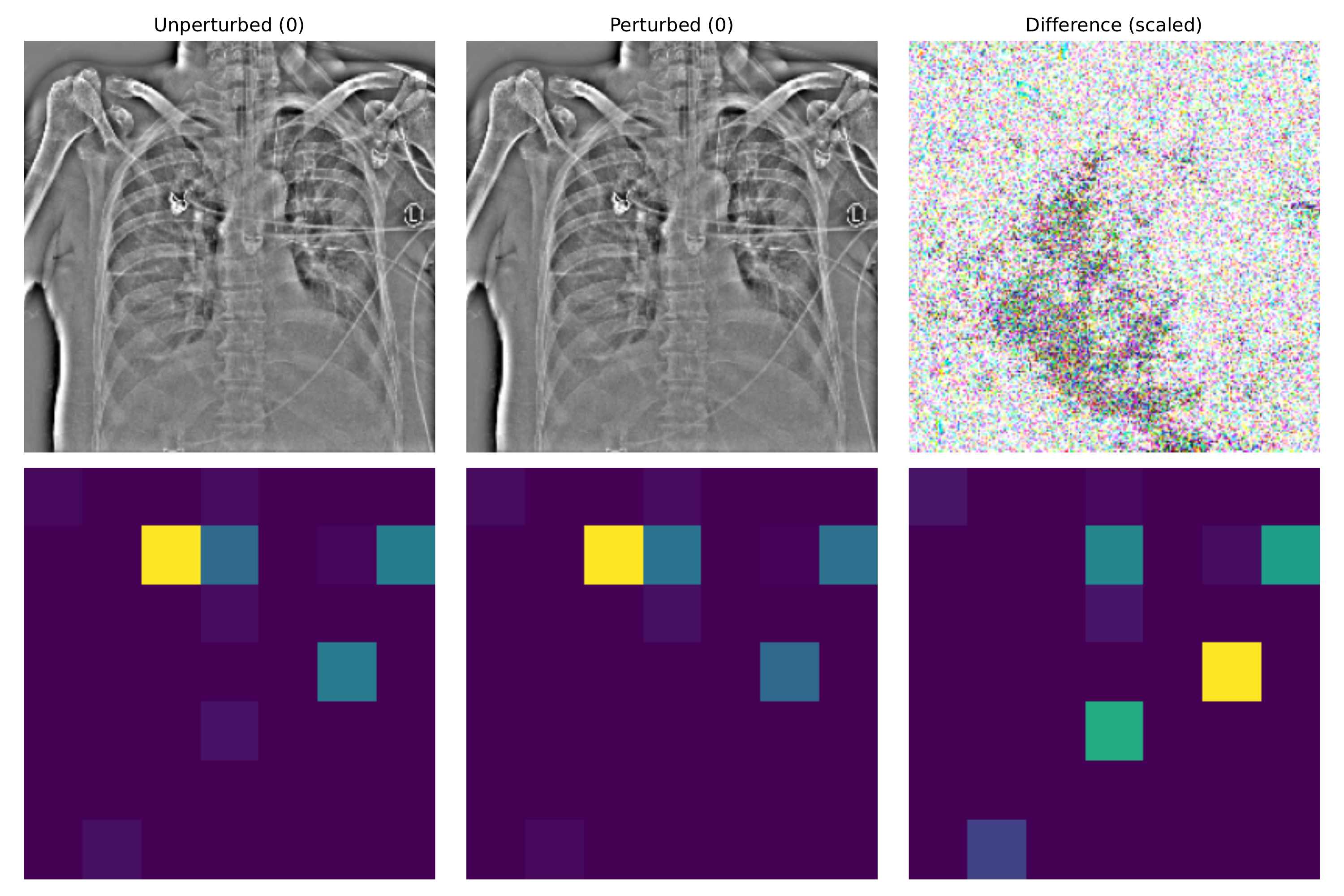}
    \caption{Comparison of activation maps (with attention) for perturbed and unperturbed chest x-ray}
    \label{PneumoWithAttention}
  \end{figure}
      \begin{figure}[h]
    \centering
    \includegraphics[width=\linewidth]{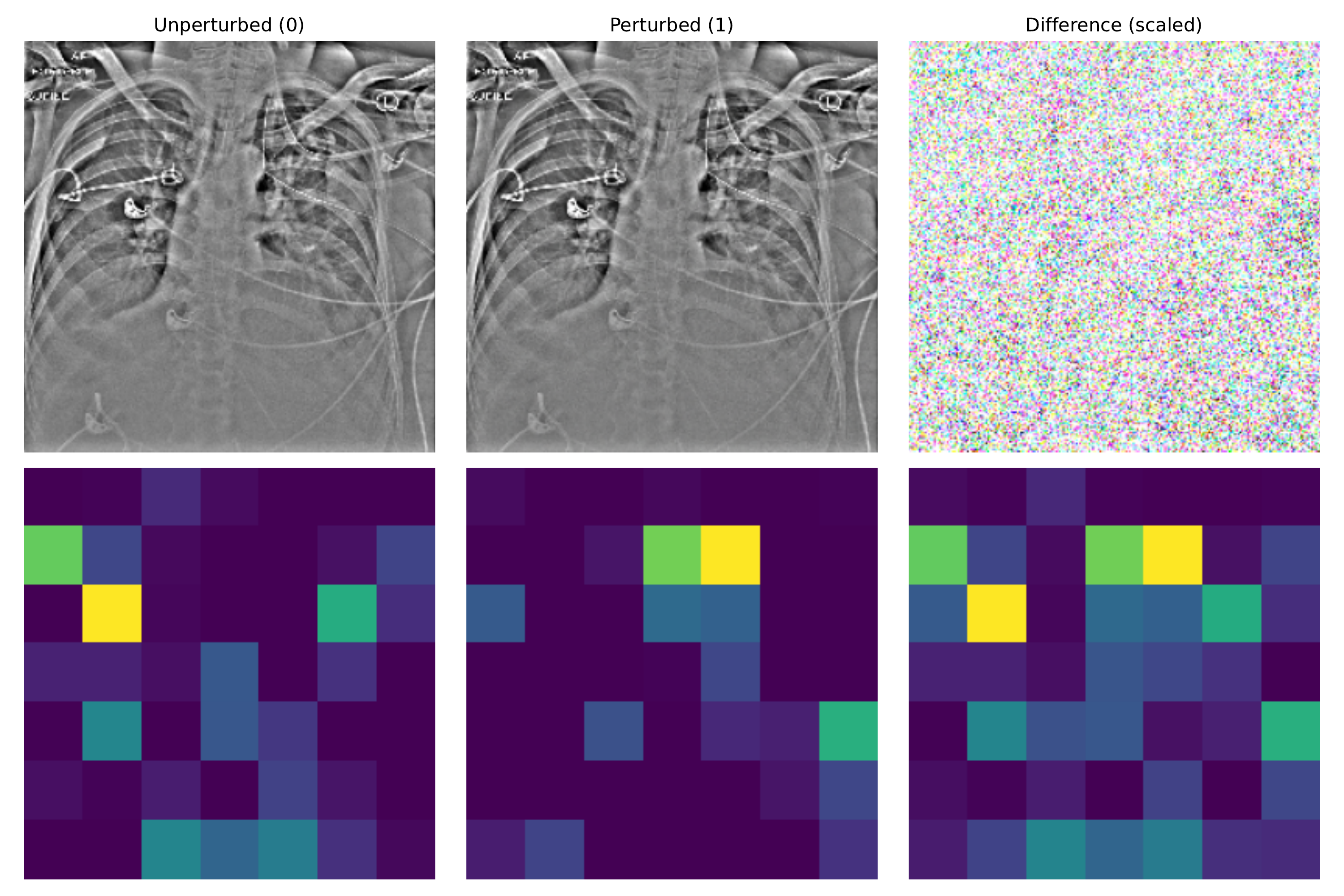}
    \caption{Comparison of activation maps (without attention) for perturbed and unperturbed chest x-ray}
    \label{DifferentMaps}
  \end{figure}

\begin{figure}[h]
    \centering
    \includegraphics[width=0.5\linewidth]{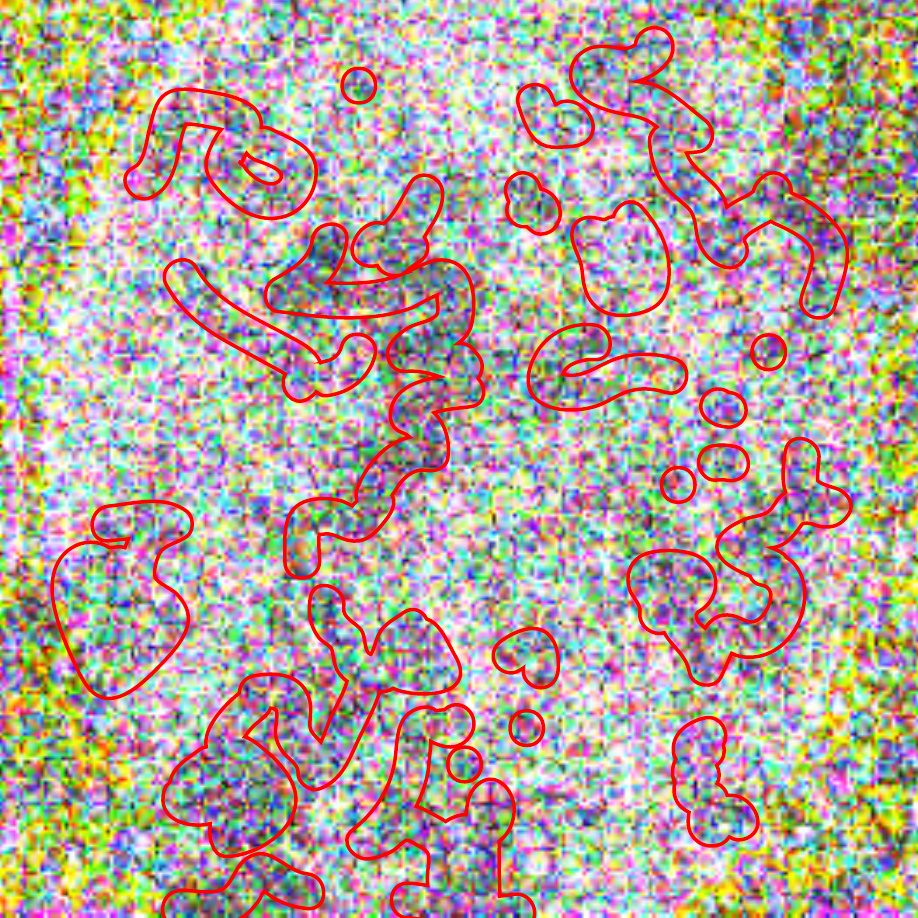}
    \caption{Dark splotches in perturbation of fundoscopic images}
    \label{CottonWool}
  \end{figure}

      In the attacks tailored to the fundoscopic image models, perturbations introduced dark splotchy patterns to the eye area  (Fig. \ref{CottonWool}). Based on diagnostic literature, these spots could be ``attempts'' by the attack to emulate cotton-wool spots or hemorrhages, key indicators of diabetic retinopathy \cite{WillsEye}. Similar patterns appear in the perturbations for the chest x-ray model with attention (Fig. \ref{PneumoWithAttention}) (these spots are larger and less speckled), though these do not appear to be similar to the key indicators of pneumothoraces (air in pleural space, misplaced lung edge, less distinct lung markings, etc.) \cite{UnofficialGuide}, it remains unclear whether the attack could be emulating these features indirectly. If so, this behavior could be responsible for medical adversarial images being so easily detected, as observed by Ma and Niu et al. (2021) \cite{MaNiu}.

      While the activation maps for the dermatoscopic and fundoscopic image models were invariant to perturbation, the chest x-ray model without attention occasionally produced very different activation maps under adversarial attacks (Fig. \ref{DifferentMaps}). However, this was only true under successful attacks.

      For all three datasets, models with attention produced activation maps that were very different from their baseline counterparts; in most cases, the regions of highest activation did not match across models. Additionally, the activation maps for the chest x-ray model with attention were highly focused, typically having a single region of significance (Fig. \ref{DifferentMaps}).

      \subsubsection{InceptionResNetV2 Models}
        Similar to the ResNet-50-based models, perturbations of dermatoscopic and fundoscopic images carried perceptible information about the source image. Across all three datasets, the perturbations for attentive models were generally higher contrast; in other words, the difference in the level of perturbation between the most- and least-perturbed pixels was greater. Like the ResNet-50 models, the perturbations targeting fundoscopic and chest x-ray images produced intensely perturbed splotches. However, unlike the ResNet-50 models, the splotches on the x-rays appear in both the baseline and attentive models and are less widespread (Fig. \ref{IRV2Comparison}).

        The activation maps for all skin lesion classification and diabetic retinopathy detection models were greatly impacted by perturbation, often being nearly inverted, indicating that the model is focusing on the incorrect regions of the image. Perturbations appeared to have less of an effect on the activation maps of the pneumothorax detection models. Like the ResNet-50 models, the activation maps of the attentive models shared no similarity with those of the baselines. Also, similar to its ResNet-50 counterpart, the attention maps of the chest x-ray models were significantly more focused than those of the other models (Fig. \ref{IRV2Comparison}).

    \subsection{Conclusion} 
      The inclusion of a soft-attention block was able to improve the robustness of ResNet-50 on two of the three medical classification tasks tested, while only slightly decreasing clean-accuracy. Additionally, previous experiments have shown the ability of this architectural modification to increase clean accuracy \cite{AttentionSkinCancerClassification}. Though soft-attention was unable to significantly improve the robust accuracy of the InceptionResNetV2 models, its inclusion was not detrimental to clean- or robust-accuracy in two of three tasks.

      These results suggest that, in most cases, the inclusion of a soft-attention block is beneficial (or at least not harmful) to the overall performance of medical image classification architectures. This modification has potential to improve accuracy and even greater potential to improve model robustness.

      Additionally, the observations above regarding perturbation behavior suggest that PGD may be inadvertently emulating cotton-wool spots, hemorrhages, and lung boundaries. The fact that these patterns are so easily visible in the perturbations may lend further insight into why adversarial images targeting medical classifiers are so easily detected.

      Future research into the susceptibility of medical image classifiers to adversarial images may wish to investigate this hypothesis by creating a modified PGD attack algorithm which maximizes the perceived randomness of the perturbation, as this strategy could lead to less detectable attacks.
    
{\small
\bibliographystyle{ieee_fullname}
\bibliography{egbib}
}

\clearpage
\appendix
  \section{Additional Figures}
    \begin{figure}[h!]
      \includegraphics[height=6in]{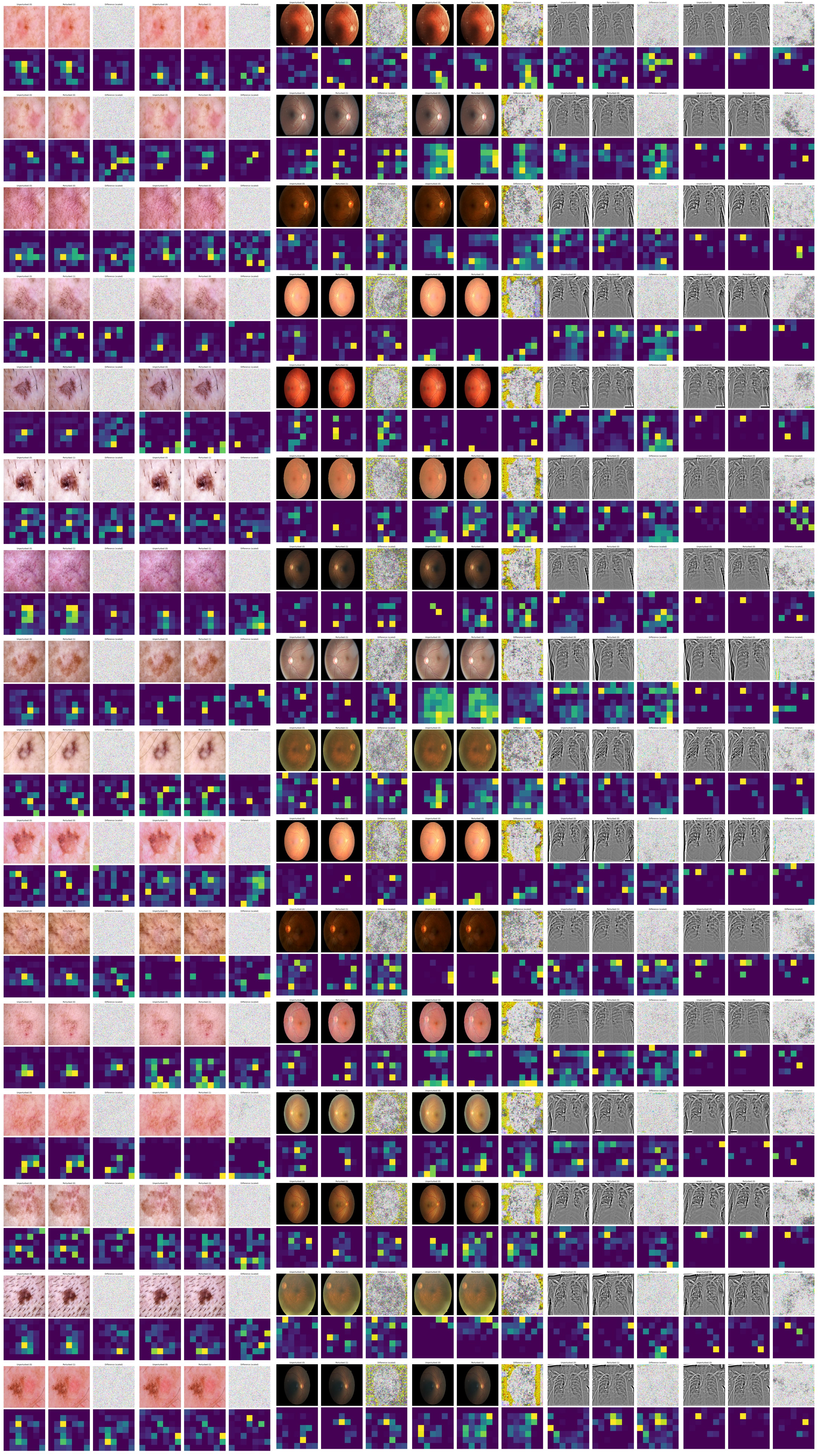}
      \caption{Perturbations and attention maps of skin lesion classification, diabetic retinopathy detection, and pneumothorax detection models using ResNet-50 architecture at $\epsilon=0.01$, $\epsilon=0.32$, and $\epsilon=0.32$ respectively.}
      \label{ResNet50Comparison}
    \end{figure}
    \clearpage
    \begin{figure}[h!]
      \includegraphics[height=6in]{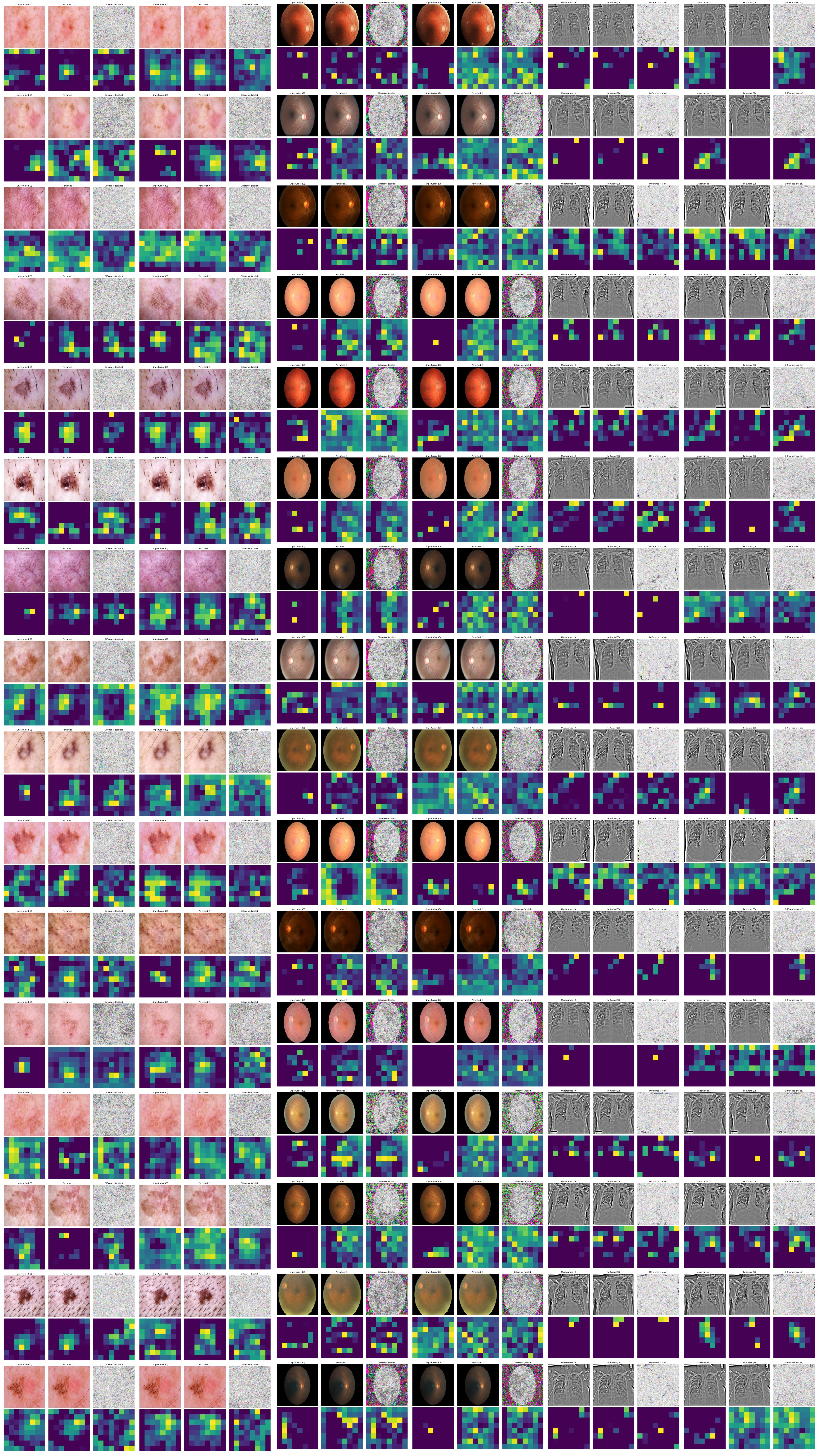}
      \caption{Perturbations and attention maps of skin lesion classification, diabetic retinopathy detection, and pneumothorax detection models using InceptionResNetV2 architecture at $\epsilon=0.00125$.}
      \label{IRV2Comparison}
    \end{figure}

\end{document}